\documentclass[runningheads]{llncs}
\usepackage{graphicx}
\usepackage{color}
\usepackage{booktabs}
\usepackage{multirow}
\usepackage{amssymb}
\usepackage{amsmath}
\usepackage{makecell}
\usepackage{booktabs}
\usepackage{longtable}
\usepackage[table,xcdraw]{xcolor}
\usepackage[colorlinks]{hyperref}

\begin{document}
\title{\texttt{TumorCP}: A Simple but Effective Object-Level Data Augmentation for Tumor Segmentation}
\titlerunning{\texttt{TumorCP}: A Simple but Effective Augmentation for Tumor Segmentation}
\author{Jiawei Yang\inst{1, \star} \and Yao Zhang\inst{2,3,\star} \and Yuan Liang\inst{1} \and Yang Zhang\inst{5} \and Lei He\inst{1,\dag} \and Zhiqiang He\inst{5,\dag}}
% index{Yang, Jiawei}
% index{Zhang, Yao}
% index{Liang, Yuan}
% index{Zhang, Yang}
% index{He, Lei}
% index{He, Zhiqiang}

\authorrunning{Jiawei Yang, Yao Zhang et al.}
\institute{Electrical and Computer Engineering, University of California, Los Angeles, USA \and Institute of Computing Technology, Chinese Academy of Sciences, Beijing, China \and University of Chinese Academy of Sciences, Beijing, China \and Lenovo Corporate Research \& Development, Lenovo Ltd., Beijing, China}

\renewcommand{\thefootnote}{}
\maketitle              % typeset the header of the contribution

\footnote{$^{\star}$: Equal contribution. ${\dag}$: Equal contribution as the corresponding authors.} 

% \author{Anonymous}

% %
% \authorrunning{Anonymous}

% \institute{}

% \maketitle              % typeset the header of the contribution
%
\begin{abstract}
Deep learning models are notoriously data-hungry. 
Thus, there is an urging need for data-efficient techniques in medical image analysis, where well-annotated data are costly and time consuming to collect. 
Motivated by the recently revived ``Copy-Paste'' augmentation, we propose \texttt{TumorCP}, a simple but effective object-level data augmentation method tailored for tumor segmentation.
\texttt{TumorCP} is online and stochastic, providing unlimited augmentation possibilities for tumors' subjects, locations, appearances, as well as morphologies. 
% We empirically show that the possible artifacts generated by Copy-Paste do not impede the overall effectiveness of \texttt{TumorCP}.
% On kidney tumor segmentation task (KiTS19 dataset), by simply copying randomly-chosen tumor from one patient and pasting it onto another patient's kidney along with straightforward object-level augmentations, \texttt{TumorCP} can surpass the strong baseline by a remarkable margin of \textbf{7.12\%} tumor Dice score.
Experiments on kidney tumor segmentation task demonstrate that \texttt{TumorCP} surpasses the strong baseline by a remarkable margin of 7.12\% on tumor Dice. Moreover, together with image-level data augmentation, it beats the current state-of-the-art by 2.32\% on tumor Dice. 
Comprehensive ablation studies are performed to validate the effectiveness of \texttt{TumorCP}. Meanwhile, we show that \texttt{TumorCP} can lead to striking improvements in extremely low-data regimes. Evaluated with only 10\% labeled data, \texttt{TumorCP} significantly boosts tumor Dice by \textbf{21.87\%}. 
%Our implementation will be made available once the paper is accepted. 
% baseline model trained with \texttt{TumorCP} augmentation with image-level augmentation on only \textbf{10\%} labeled data can have \textbf{21.87\%} tumor Dice improvement. 
%We see the success of \texttt{TumorCP} as a favor of context-decoupled learning in medical images realm, and summarize its effectiveness from three perspectives: i) eliminating background bias by context-invariant prediction, ii) improving generalizability by transformation-invariant prediction, and iii) oversampling behaviors. 
% More importantly, we show that \texttt{TumorCP} can lead to striking improvements on the extreme low-data regime. 
To the best of our knowledge, this is the very first work exploring and extending the ``Copy-Paste'' design in medical imaging domain. 
Code is available at: \href{https://github.com/YaoZhang93/TumorCP}{https://github.com/YaoZhang93/TumorCP}.
%Our implementation will be made available once the paper is accepted. 

\end{abstract}
\section{Introduction}

Deep learning (DL) models work remarkably well over the past few years in computer vision tasks, including medical image analysis. Though DL models act like de facto standard, they are notoriously data-hungry, demanding more so than ever large and well-annotated datasets to achieve robust performance \cite{xie2020self}. However, high-quality annotated datasets require intense labor and domain knowledge, which becomes more expensive in the medical domain. 

To improve data-efficient learning, several successful approaches have been proposed from different perspectives, such as leveraging unlabeled data for semi-supervised self-training \cite{xie2020self,chen2020simple,sohn2020fixmatch} or self-supervised pre-training \cite{zhou2019models,chen2020simple,mitrovic2020representation}, distilling priors from data as explicit constraints for model training~\cite{liang2020atlas,liang2019comparenet}, generating new data with the imaging of an anatomy of a different modality~\cite{liang2021oralviewer,liang2020x2teeth}, or utilizing appropriate data augmentation methods to increase data diversity \cite{ghiasi2020simple,shorten2019survey,dwibedi2017cut,sohn2020fixmatch}. Some of them are designated for medical images. Particularly, Zhou et al. \cite{zhou2019models} designed a unified self-supervised learning framework, integrating multiple proxy tasks to exploit unlabeled medical data, and showed performance gains for downstream tasks. Xue et al. \cite{xue2019synthetic}, and Shin et al. \cite{shin2018medical} used GANs to generate additional training data for histopathology image classification and brain tumor segmentation. The quality of the ``realness'' of synthesized training data dramatically affects model performance due to the risk of overfitting to fake data. Eaton et.al. \cite{eaton2018improving} studied Mix-up \cite{zhang2017mixup} augmentation for brain tumor segmentation. However, it requires a specific patch-level operation which involves complicated strategies, e.g. sampling of small patches to be mixed up.

% Not following the way, We tend to break the trend of using increasingly sophisticated art methods like GANs. 
Distinct from the trend of using increasingly sophisticated methods like GANs, we investigate ``Copy-Paste'', a straightforward augmentation technique \cite{fang2019instaboost,dwibedi2017cut} that has been recently revisited and made breakthroughs in natural image instance segmentation \cite{ghiasi2020simple}. 
Copy-Paste augmentation avoids costly generation processes from representation space to pixel space by simply pasting the labeled instance onto new background images as additional training data.  
% and can achieve better performance \cite{ghiasi2020simple}. 
Despite its success in natural images, such method is largely unexplored in the medical image realm. 
Moreover, its effectiveness for medical tasks remains doubtable since the context information tends to be ignored in Copy-Paste. 
For instance, in the tumor segmentation, one would argue the importance of surrounding visual clues, i.e., context, for the emergence of a tumor. 
Besides, one would believe the inherent anatomical structures in medical image make the context indispensable for tumor segmentation. 
% The ``context-aware'' topic keeps its popularity in the medical field with few protesters.
In this work, we also aim to fill the gap of understanding the role of \textit{context} in medical domain by examining the effectiveness of Copy-Paste augmentation for tumor segmentation. 
% We show that the context-decoupled method can improve data-efficiency dramatically without loss of generality. 

We propose \texttt{TumorCP}, a simple but effective object-level data augmentation method based on Copy-Paste for tumor segmentation tasks. Straightforwardly, \texttt{TumorCP} randomly chooses a tumor from a source image and paste it onto the organs in the target image after a series of spatial, contrast, and blurring augmentations. We use kidney tumor segmentation (KiTS19 dataset \cite{heller2019kits19}) and a state-of-the-art model (\texttt{nnUNet}\cite{isensee2021nnu}) as the benchmark to evaluate the proposed method. We empirically show that though \texttt{TumorCP} inevitably generates artifacts after Copy-Paste, it consistently provides solid gains over all different settings in our experiments. Specifically, with only rigid spatial transformation and Copy-Paste within the same patient, \texttt{TumorCP} can surpass the baseline by 6.24\% tumor Dice. Together with inter-patient Copy-Paste and other tumor-oriented augmentations, \texttt{TumorCP} further outperforms the baseline by 7.12\% tumor Dice. Moreover, with image-level data augmentation (ImgDA), our best version beats state-of-the-art by 2.3\%. Going one step further, we also study \texttt{TumorCP} for extremely low-data regime, where only 10\% labeled data are exploited for training. Under this setting, \texttt{TumorCP} with ImgDA can improve the tumor Dice by \textbf{21.87\%} compared with no-data-augmentation (noDA), which 
% to the best of our knowledge, 
is unprecedented to our knowledge, convincingly demonstrating the effectiveness of \texttt{TumorCP} for data-efficiency learning. 

The success of \texttt{TumorCP} is an empirical observation to support context-decoupled learning even in \textit{medical domain}. We briefly discuss our understanding of the open question of the context's role and why \texttt{TumorCP} works in Sec \ref{sec:discussion}. We hope our work can provide some useful data points to our community and shed light on the importance of Copy-Paste augmentation, which is powerful but unfortunately nearly absent in the medical imaging field.

% \vspace{-5mm}
\section{Method}
% \vspace{-2mm}

\begin{figure}[h]
\includegraphics[width=\textwidth,height=6.4cm]{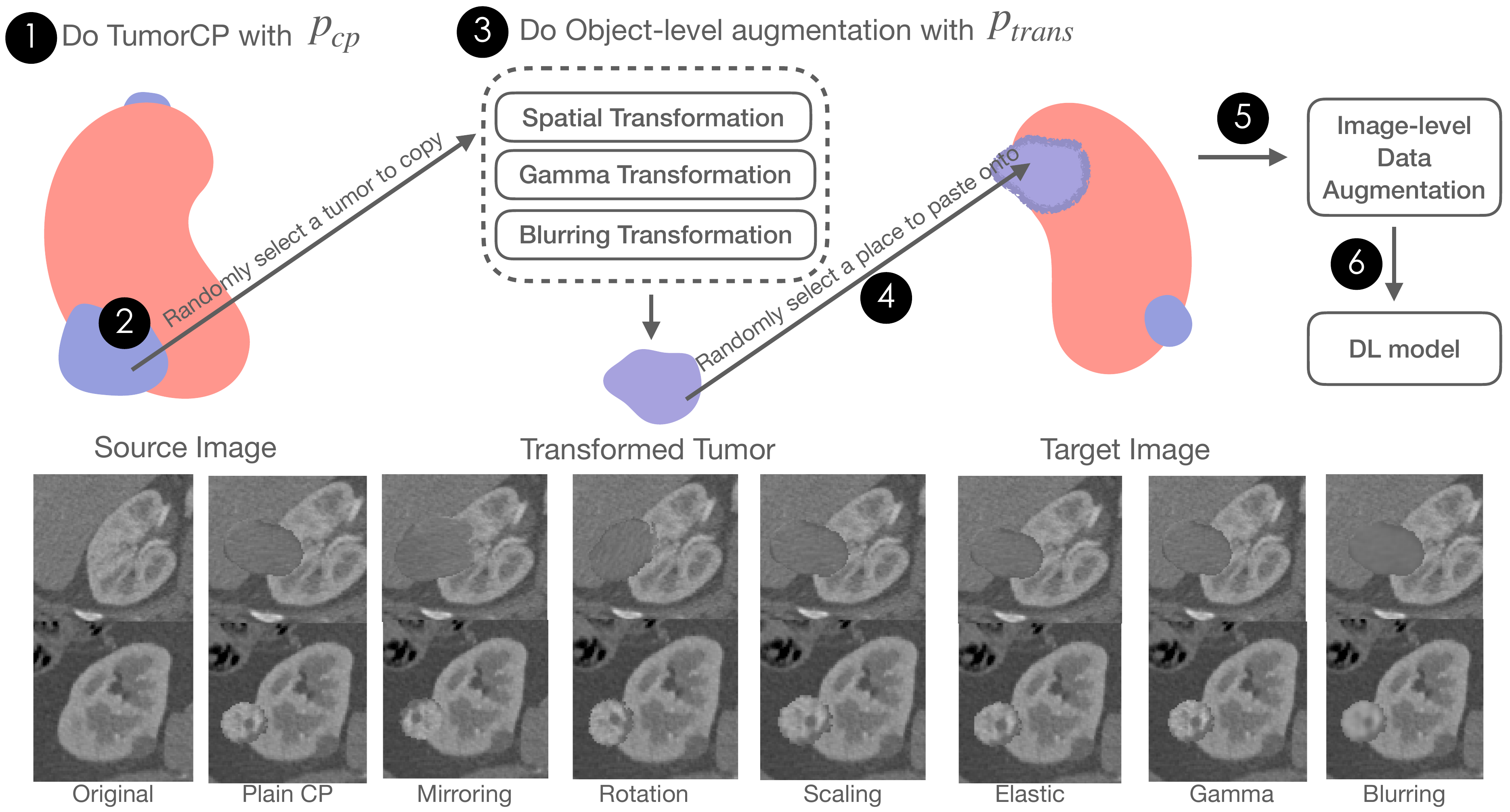}
\caption{Illustration of \texttt{TumorCP}'s pipeline. A pair of source image and target image are sampled from the dataset. With probability of $p_{cp}$, \texttt{TumorCP} performs Copy-Paste once, following the step number of 2,3,4,5, and finally to 6; otherwise, it directly goes to step 6. In step 3, each of the transformation has its own probability ($p_{trans}$) to be invoked. The bottom illustrates two samples performing Copy-Paste with object-level data augmentation.}
\label{fig:tumorcp_overview}
\end{figure}

\texttt{TumorCP} is an online and stochastic augmentation process specified for tumor segmentation. Its implementation is easy and straightforward. As illustrated in Fig.\ref{fig:tumorcp_overview}, given a set of training samples $\mathcal{D}$, with the probability of $(1-p_{cp})$, \texttt{TumorCP} does nothing; otherwise \texttt{TumorCP} samples a pair of images $(x_{src}, x_{tgt}) \sim \mathcal{D}$ and conducts Copy-Paste once. Let $\mathcal{O}_{src}$ be the set of tumor(s) on $x_{src}$, $\mathcal{V}_{tgt}$ be the set of volumetric coordinates of organ(s) on $x_{tgt}$, and $\mathcal{T}$ be the set of stochastic data transformations, each of which has a probability parameter called $p_{trans}$. To do once Copy-Paste, \texttt{TumorCP} first samples a tumor $o \sim \mathcal{O}_{src}$, a set of transformation(s) $\tau \sim \mathcal{T}$, and a target location $v \sim \mathcal{V}_{tgt}$, followed by centering $\tau(o)$ at $v$ to replace the original data and annotation.  To fully leverage the advantage of \texttt{TumorCP}, we carefully design two modes of Copy-Paste for tumors: intra-patient and inter-patient Copy-Paste. Meanwhile, we enhance Copy-Paste with several object-level transformation to obtain abundant augmentations.
% \vspace{-0.5cm}
% \begin{figure}[!tp]
% \centering
% \includegraphics[width=\linewidth]{figs/examples.pdf}	
% \caption{Illustration of CT images augmented by \texttt{TumorCP}. \texttt{TumorCP} performs an object-level augmentation, copying tumors to another location within a CT image or to another CT image. Furthermore, \texttt{TumorCP} enables object-level transformation to achieve abundant augmentation.}
% \label{examples}
% \end{figure}
\subsection{\texttt{TumorCP}'s augmentation}
\subsubsection{Intra-/Inter- Copy-Paste.} \label{sec:consider}
 In order to study the effect of inter-patient variance to \texttt{TumorCP}, we define two base settings: 1) intra-patient Copy-Paste (intra-CP) if the source and target images are identical, i.e., both from the same patient and 2) inter-patient Copy-Paste (inter-CP) if those are different. From the perspective of data distribution, the intra-CP is preferred as its intensity agreement with the data as a whole, but this limits data diversity. From the perspective of data diversity, the inter-CP is favored as it unlocks the access for leveraging both new backgrounds and foregrounds from other patients, but it also brings distribution discrepancy. It might be surprising that we empirically show the inter-CP significantly outperforms intra-CP one in ablation study in Sec \ref{sec:inter_ablation}.
% \vspace{-0.5cm}
%we define two scenarios: i) intra-subject copy-paste (intraCP) and ii) inter-subject CP (interCP). For intraCP, we randomly select a tumor from one subject and paste it onto the same subject's kidney. For interCP, in contrast, two subjects are randomly chosen; tumor(s) selected from one subject is pasted onto the other's kidney. The segmentation labels of the pasted-onto subject are replaced accordingly. More details can be found in {\color{red}Appendix.} It is worthy of mentioning that all the instance augmentation process is both \textbf{online} and \textbf{random}, bringing unlimited possibilities for tumors' locations and appearances within or across the subjects.

\subsubsection{Copy-Paste with Transformations.} Building from plain Copy-Paste, we naturally extend it by incorporating four different object-level transformations motivated by different objectives as the followings. The detailed implementations are summarized in appendix. 
\let\labelitemi\labelitemii
\begin{itemize}
	\item \textbf{Spatial transformation decouples context and improves morphology diversity.} Given the \textit{fixed} acquired CT images, tumors always appear along with their surrounding visual context. Though image-level spatial augmentation increases data diversity in terms of perspectives (e.g., mirroring and slight rotation), it still processes an image as a whole, remaining the coupling between foreground and background. Therefore, the model can seek for and tend to overfit to the plausible but de facto irrelevant surrounding clues. Note that plain Copy-Paste already addressed this problem by offering new background via the most basic spatial transformation --- \textbf{shifting}. We further increase the morphology diversity by applying i) rigid transformation that includes scaling, rotation, and mirroring, and ii) elastic transformation that deforms tumors. Fig. 1 demonstrates examples of transformed tumors.
	\item \textbf{Gamma transformation enhances contrast and improves intensity diversity.} Given a tumor, we apply gamma transformation to adjust its intensity distribution while retaining the whole intensity range. On the one hand, the tumor intensity diversity is enhanced by randomly sample gamma parameter; on the other hand, the local contrast is enhanced by power-law non-linearity, facilitating tumor discrimination. 
	\item \textbf{Blurring transformation improves texture diversity.} We use a Gaussian filter as the blurring transformation. Intuitively, a Gaussian filter with different sigma values can filter out the noise and smooth the tumor to some extent. Aggregating noise-perturbed low-level textures can indirectly increase the texture diversity to relatively high-level textures. 
\end{itemize}

The whole pipeline can be incorporated together with image-level augmentation. It is worth mentioning that all the instance augmentation process is both \textbf{online} and \textbf{random}, bringing unlimited possibilities for tumors' locations and appearances within or across the subjects.

\subsection{Intuitions on  \texttt{TumorCP}'s Effectiveness} \label{sec:discussion}

As aforementioned, \texttt{TumorCP} has two goals: i) increase the data diversity, and ii) learn high-level and to abstract the \textit{invariant} representation of tumor. Data diversity is increased as the new combinations of tumors, and their surroundings are generated with the augmentation. For learning high-level information, we discuss three properties of \texttt{TumorCP} to explain its effectiveness.\\
\textbf{Eliminated background bias by context-invariant prediction.} As mentioned before, the semantic contexts are \textit{fixed} for the acquired medical images. Convolutional Neural Network (CNN) inevitably convolutes surrounding visual contexts along with the objects themselves. This can bias the model towards plausible but indeed tumor-irrelevant clues, increasing the risk of overfitting. With both \textit{\textbf{random}} and \textit{\textbf{online}} spatial transformation, \texttt{TumorCP} offers access for tumor to preciously unattached zones and thus provides unlimited possibilities for tumors' surrounding contexts. It enforces the model's prediction to be invariant across different visual surroundings and eliminates background bias.\\
\textbf{Improved generalizability by transformation-invariant prediction.} The model should capture both high-level semantic information and low-level boundary information for successful segmentation. With both \textit{\textbf{random}} and \textit{\textbf{online}} Gamma \& Blur transformations, \texttt{TumorCP} can generate diverse tumors in terms of size, shape, color and texture, which increase the intra-class disparity. It tasks the model to capture the golden semantics from the data. In other words, it enforces the model's prediction to be invariant across different data transformation (that potentially resembles real-world data) and improves generalizability.\\
\textbf{Oversampling behavior.} Data imbalance is a widely experienced problem. Typical solutions usually re-weight loss function or re-sample training data according to the class distribution. In this work, the distribution of background, organ, and tumor is extremely imbalanced. From this perspective, \texttt{TumorCP} acts like a data re-sampler that significantly increases the volume of tumors in multiplication degree at a minor cost. 

\section{Experiments and Discussion}
\subsection{Experiment settings} 

We evaluate \texttt{TumorCP} on KiTS19 \cite{heller2019kits19}, a publicly available dataset for kidney tumor segmentation. We randomly split the published 210 images into a training set with 168 images and a validation set with 42 images. As the limited computation resources, we majorly report ablation study results on the validation set if not specified. Note that this validation set is unaugmented and unseen i.e., neither used to tune hyper-parameters nor to monitor the training process. 
We use S{\o}rensen-Dice Coefficient (Dice) score in all experiments, 
which measures the overlap of model's prediction $y_{pred}$ and ground truth $y_{true}$, formulated as $\mathrm{Dice}=|y_{true}\cap y_{pred}|/|y_{true} \cup y_{pred}|$. The average and standard deviation of the Dice score over all patients are reported.
% We study the proposed \texttt{TumorCP} on KiTS19 dataset \cite{heller2019kits19}, a publicly available dataset for kidney tumor segmentation which has 300 CT images with both kidneys and kidney tumors annotated. The organizers keep 90 images unreleased for testing. For data splitting, we randomly divide training data into 5-fold. Due to limited computation resources, we report our major ablation study results on one validation fold if not specified. For all experiments, we use Dice-S{\o}rensen Coefficient (Dice) score as our metric.

We use publicly available\footnote{https://github.com/MIC-DKFZ/nnUNet} state-of-the-art \texttt{nnUNet} codebase for implementation, which includes data pre-processing, leading image-level augmentation pipelines, as well as top-performance models. It almost tops all biomedical image segmentation benchmarks \cite{isensee2021nnu}. This paper focuses on a general augmentation method for tumor segmentation, so the choices of datasets and running models are orthogonal to our goal. \texttt{TumorCP} can generalize to other segmentation models and tumor segmentation datasets at no cost. 

All experiments are conducted on Nvidia V100 GPU with 500 epochs training of \texttt{3d\_fullres nnUNet}, instead of 1000 epochs by nnUNet's default. The batch size for training is 2. During training, each epoch takes 250 iterations, which means 250 batches of data are sampled and learned. Other settings in model training remain its default. We refer readers to \cite{isensee2021nnu} and the codebase link for more details.

\subsection{Ablation Study}

For simplicity and unification, we set the probability of \texttt{TumorCP} performing Copy-Paste as $p_{cp}=0.8$ for all experiments.\\

\textbf{Ablation on intra-CP with different transformations.} We first investigate \texttt{TumorCP} under intra-CP with various object-level transformations. In this ablation, no image-level augmentation is applied. All object-level transformations have a 0.5 probability of being invoked. For example, Intra-CP\&Rigid means rigid transformation has a 0.5 probability to be conducted when Intra-CP is triggered. Table \ref{tab:ablation1} presents the comparison on different methods. As the first group of Table \ref{tab:ablation1} demonstrates, all the models trained with \texttt{TumorCP} (shaded cells) consistently outperform the baseline model, no-data-augmentation (noDA). Specifically, the vanilla intra-CP itself can bring 1.09\% Dice improvement over baseline; \texttt{TumorCP} with only rigid transformation can increase tumor Dice by 6.24\%.

%Remarkably, composited with all transformations, the intra-CP based \texttt{TumorCP} boost the baseline model by a significant value of {\color{red}\textbf{8.56\%}}, which firmly confirms the effectiveness of \texttt{TumorCP}.
% \vspace{-5mm}
\begin{table}[]
\centering
\caption{Ablation study of \texttt{TumorCP}. The first group shows the results of \texttt{TumorCP} with different transformations in the intra-CP setting, while the second group shows the results \texttt{TumorCP} with intra-CP and inter-CP settings. The shaded rows denote our work.}
\label{tab:ablation1}

\setlength{\tabcolsep}{6mm}{
\resizebox{1.0\textwidth}{!}{
\begin{tabular}{c|c|c} 
\toprule
\multicolumn{1}{c|}{\multirow{2}{*}{\textbf{Method}}}    & \multicolumn{2}{c}{\textbf{Mean Dice $\pm$ std / improvement over baseline} (\%) $\uparrow $}                                                               \\  \cmidrule(l){2-3}
\multicolumn{1}{c|}{}     & Kidney        & \multicolumn{1}{c}{Tumor}             \\ \midrule\midrule
							noDA                 & 96.62$\pm$2.41/baseline                       & 72.59$\pm$26.97/baseline                                        \\ \midrule
\rowcolor[HTML]{EFEFEF} Intra-CP                 & 96.81$\pm$2.02/+0.19                 & 73.68$\pm$26.99/+1.09                               \\
\rowcolor[HTML]{EFEFEF} Intra-CP\&Elastic        & 96.75$\pm$1.88/+0.13                 & 73.95$\pm$28.20/+1.36                              \\
\rowcolor[HTML]{EFEFEF} \textbf{Intra-CP\&Rigid} & 96.78$\pm$1.92/+0.16                 & \textbf{78.83$\pm$19.77/+6.24}             \\
\rowcolor[HTML]{EFEFEF} Intra-CP\&Gamma          & 96.81$\pm$1.89/+0.19                 & 76.32$\pm$23.97/+3.73                               \\
\rowcolor[HTML]{EFEFEF} Intra-CP\&Blur           & \textbf{96.89$\pm$1.92/+0.27}        & 76.46$\pm$24.86/+3.87                         \\ \midrule
\multicolumn{1}{l}{}                                            & \multicolumn{1}{l}{}                                                & \multicolumn{1}{l}{}      \\
\midrule\midrule
\rowcolor[HTML]{EFEFEF} Intra-CP & \textbf{96.81$\pm$2.02/+0.19}  & 73.68$\pm$26.99/+1.09   \\
\rowcolor[HTML]{EFEFEF} Inter-CP & 96.73$\pm$2.03/+0.11  & 77.22$\pm$23.67/+4.63\\
\rowcolor[HTML]{EFEFEF} \textbf{Intra-\&Inter-CP} & 96.78$\pm$1.98/+0.16  & \textbf{77.44$\pm$23.46/+4.85}\\
\bottomrule
\end{tabular}%
}}
\end{table}

\subsubsection{Ablation on intra-/inter-CP.}\label{sec:inter_ablation} Here we study the effect of intra-/inter-CP for the considerations in Sec. \ref{sec:consider}. The second group in Table 1 shows that inter-CP significantly outperforms intra-CP by 3.54\% Tumor Dice, yielding a 4.63\% improvement over the baseline model. Though surprised to some extent, this result meets our expectation as both the tumors'  and the backgrounds' diversity from one patient are still limited compared to other patients. Copying others' tumors and pasting them onto current patients' cases is supposed to unlock more novel combinations and bring more data diversity. We also aggregate intra- and inter- CP by setting a 50\% chance for each to sample data pairs from the dataset. The last line in Table \ref{tab:ablation1} presents the result and is shown to the best entry among this ablation. It demonstrates the superiority of combining both intra- and inter- patient's context exchange. 

%The results in the last and the best entry in the second group of Table 1. 

% \begin{table}
% \centering
% \caption{Ablation study of \texttt{TumorCP} with intra-CP and inter-CP. }
% \label{tab:ablation2}
% \setlength{\tabcolsep}{6mm}{
% \resizebox{0.95\textwidth}{!}{
% \begin{tabular}{c|c|c} 
% \toprule
% \multicolumn{1}{c|}{\multirow{2}{*}{\textbf{Method}}}    & \multicolumn{2}{c}{\textbf{Mean Dice $\pm$ std / improvement over baseline} (\%) $\uparrow $}                                                               \\  \cmidrule(l){2-3}
% \multicolumn{1}{c|}{}     & Kidney        & \multicolumn{1}{c}{Tumor}             \\ \midrule\midrule
% noDA & 96.62$\pm$2.41/baseline   & 72.59$\pm$26.97/baseline   \\ 
% \midrule
% \rowcolor[HTML]{EFEFEF} Intra-CP & \textbf{96.81$\pm$2.02/+0.19}  & 73.68$\pm$26.99/+1.09   \\
% \rowcolor[HTML]{EFEFEF} Inter-CP & 96.73$\pm$2.03/+0.11  & 77.22$\pm$23.67/+4.63\\
% \rowcolor[HTML]{EFEFEF} \textbf{Intra-\&Inter-CP} & 96.78$\pm$1.98/+0.16  & \textbf{77.44$\pm$23.46/+4.85}\\
% \bottomrule
% \end{tabular}
% }}
% \end{table}

\noindent\textbf{Ablation on compatibility.} As the last step, we accumulate the composition of all object-level transformations and Intra-\&Inter-CP to constitute \texttt{TumorCP}$^\star$. Previously we improve from noDA baseline. Here we also explore the compatibility between \texttt{TumorCP} and image-level augmentation. The image-level augmentation follows \texttt{nnUNetV2Trainer} default setting detailed here\footnote{https://git.io/Jqvro} \cite{isensee2021nnu}. Results in Table \ref{tab:ablation2} shows that \texttt{TumorCP$^\star$} is compatible with image-level augmentations, and thus can act as a plug-in module in general augmentation pipeline. Together with image-level augmentation, \texttt{TumorCP$^\star$} can improve 7.12\% from no image-level augmentation (noDA) baseline and 2.32\% from image-level augmentation (ImgDA) baseline. It is worthy to mention that the ImgDA baseline currently still holds the state-of-the-art performance for KiTS Dataset, which means \texttt{TumorCP$^\star$} can further boost exisiting arts to higher performance. \texttt{TumorCP$^\star$} can generalize to other models and datasets at almost no cost.

\begin{table}[t]
\centering
\caption{Comparison of \texttt{TumorCP} and image-level augmentation. The shaded rows denote our work.}
\label{tab:ablation2}
\setlength{\tabcolsep}{4mm}{
\resizebox{1.0\textwidth}{!}{
\begin{tabular}{c|c|c} 
\toprule
\multicolumn{1}{c|}{\multirow{2}{*}{\textbf{Method}}}    & \multicolumn{2}{c}{\textbf{Mean Dice $\pm$ std / improvement over baseline} (\%) $\uparrow $}                                                               \\  \cmidrule(l){2-3}
\multicolumn{1}{c|}{}     & Kidney        & \multicolumn{1}{c}{Tumor}             \\ \midrule\midrule 
											noDA 				& 96.62$\pm$2.41/baseline   	& 72.59$\pm$26.97/baseline   \\  
\rowcolor[HTML]{EFEFEF} \texttt{TumorCP}$^\star$ 				& \textbf{96.86$\pm$1.91/+0.24} & \textbf{79.71$\pm$22.56/+7.12}   \\
\midrule
													ImgDA 		& 97.06$\pm$1.48/baseline   	& 82.43$\pm$21.29/baseline   \\ 
\rowcolor[HTML]{EFEFEF} \texttt{TumorCP}$^\star$ + ImgDA 		& \textbf{97.15$\pm$1.43/+0.09} & \textbf{84.75$\pm$20.87/+2.32} \\
\bottomrule
\end{tabular}
}}
\end{table}

%\subsubsection{Comparisons with top winners in KiTS.} We perform 5-fold training using our best setting, \texttt{TumorCP$^\star$}, and use the averaged predictions from 5 models as final results. Table X shows that, we outperform the current top-1 winner by a significant margin of xx\%.

%\noindent\textbf{\texttt{TumorCP} generates artifacts but consistently has solid gains.} {\color{red} Though \texttt{TumorCP} is intended for better tumor segmentation, it also consistently improves kidney segmentation performance compared to its baselines. It also meets our intuitions for \texttt{TumorCP}, since the tumors are relative context to the kidney to some extend, which invokes \textit{``Eliminated background bias by context-invariant prediction''} but now for the kidney.}

\noindent\textbf{\texttt{TumorCP} also improves organ segmentation.} Though \texttt{TumorCP} is intended for better tumor segmentation, it also consistently improves kidney segmentation performance compared to its baselines. It also meets our intuitions for \texttt{TumorCP}, since from the perspective of kidney, tumors are the relative context and background to some extent, which resembles \textit{``Eliminated background bias by context-invariant prediction''} but now for the kidney.

\subsection{Towards extremely low-data regime} 

Finally, we demonstrate the potentials of \texttt{TumorCP} in extremely low-data regime via some additional ablations. Particularly, we randomly select 10\% data from the training set same as before. Then, we train three models, \texttt{noDA}, \texttt{ImgDA} and \texttt{TumorCP$^\star$ + \texttt{ImgDA}} on 10\% data respectively, followed by the evaluation on the same validation set. Table \ref{tab:ablation3} shows the results. Under this setting,
% \texttt{TumorCP}$^\star$ with ImgDA 
our method can improve the noDA by \textbf{21.87\%}, which, to the best of our knowledge, is \textit{unprecedented}, convincingly demonstrating the effectiveness of \texttt{TumorCP} for data-efficiency learning. It breaks the trend of using sophisticated methods or strategies while achieving promising results in low-data regime of tumor segmentation.

\begin{table}[t]
\centering
\caption{Comparison of \texttt{TumorCP} and image-level augmentation for data-efficient segmentation. The shaded rows denote our work.}
\label{tab:ablation3}
\setlength{\tabcolsep}{2mm}{
\resizebox{\textwidth}{!}{
\begin{tabular}{c|c|c} 
\toprule
\multicolumn{1}{c|}{\multirow{2}{*}{\textbf{Method}}}    & \multicolumn{2}{c}{\textbf{Mean Dice $\pm$ std / improvement over baseline} (\%) $\uparrow $}                                                               \\  \cmidrule(l){2-3}
\multicolumn{1}{c|}{}     & Kidney        & \multicolumn{1}{c}{Tumor}             \\ \midrule\midrule  
10\%-data noDA & 93.25$\pm$4.41/baseline  & 41.12$\pm$39.58/baseline  \\  
10\%-data ImgDA & 95.41$\pm$3.25/+2.16  & 54.34$\pm$31.59/+13.22  \\ 
\midrule
\rowcolor[HTML]{EFEFEF} 10\%-data  \texttt{TumorCP}$^\star$ + ImgDA& \textbf{95.53$\pm$3.25/(+2.16/+2.28)} & \textbf{62.99$\pm$26.92/(+13.22/+21.87)}\\
\bottomrule
\end{tabular}
}}
\end{table}

\section{Conclusion and Future Works}
This key contribution of our work is the proposal and comprehensive study of \texttt{TumorCP}, a simple but effective object-level data augmentation for tumor segmentation. Extensive experiments confirm the remarkable effectiveness of our method.
In addition to surpassing current art in kidney tumor segmentation by 2.31\% in tumor Dice, we also demonstrate the potential of \texttt{TumorCP} for the extremely low-data regime. We prefer to call our \texttt{TumorCP} as a \textit{new baseline}, as it does not involve any sophisticated techniques nor extensive hyper-parameter adjustment while achieving the new state-of-the-art. 
Besides, \texttt{TumorCP} does not directly handle the distribution mismatching in the inter-CP setting but still gets fabulous performance. 
% Perhaps with designated processing, \texttt{TumorCP} can surprise us once more. 
% Future works can easily extend \texttt{TumorCP} for other medical segmentation tasks, such as \texttt{OrganCP} for anatomical segmentation and \texttt{TissueCP} for . 
Future works can easily extend \texttt{TumorCP} for other medical segmentation tasks without significant modifications, and are worth trying for further improving state-of-the-art accuracy.

\bibliographystyle{splncs04}
\bibliography{paper1885.bib}

\appendix
\section{Appendix -- Data augmentation details}
\subsection{Rigid Transformation}
Rigid transformation consists of three transformation:
\begin{itemize}
	\item Mirroring: randomly choose one from eight possible mirroring axes combination, i.e. (x), (y), (z), (x,y), (x,z), (y,z), (x,y,z), with probability of 0.5.
	\item Rotation: Since the 3D abdominal CT data are usually anisotropic. We only rotate the instance around z-axis to constrain the spacing consistency. The tumor will rotate randomly in a range of $(-\pi, \pi)$ with the probability of 0.5.
	\item Scaling: With the probability of 0.5, the tumor will be re-scaled in a range of $(0.75, 1.25)$ with \texttt{resize} function of \texttt{skimage} package in an order 3. 
\end{itemize}

\subsection{Deformable Elastic Transformation} We use the implementation of \texttt{batchgenerators}\footnote{Publicly available at: https://git.io/JqfTt} python library, with the alpha range $(0,900)$ and sigma range $(9,13)$.

\subsection{Gamma Transformation}
For gamma transformation, we force the mean and standard deviation of the copied instance unchanged. The gamma value is sampled in a range of (0.7, 1.5).

\subsection{Blurring Transformation}
We use Gaussian filter with the sigma sampled from $(0.5, 1)$ as blurring transformation.

\end{document}